\documentclass{article}

\usepackage[final]{nips_2017}

\usepackage[utf8]{inputenc} 
\usepackage[T1]{fontenc}    
\usepackage{hyperref}       
\usepackage{url}            
\usepackage{booktabs}       
\usepackage{amsfonts}       
\usepackage{nicefrac}       
\usepackage{microtype}      
\usepackage{graphicx}
\usepackage{subfig}
\usepackage{color}

\title{Lip2AudSpec: Speech reconstruction from silent lip movements video}

\author{
  Hassan Akbari \\
  Department of Electrical Engineering \\
  Columbia University, New York, NY, USA \\
  \texttt{ha2436@columbia.edu} \\
\And
  Himani Arora \\
  Department of Electrical Engineering \\
  Columbia University, New York, NY, USA \\
  \texttt{ha2434@columbia.edu} \\
\And
  Liangliang Cao \\
  Department of Electrical Engineering \\
  Columbia University, New York, NY, USA \\
  \texttt{lc2922@columbia.edu} \\
 \And
  Nima Mesgarani \\
  Department of Electrical Engineering \\
  Columbia University, New York, NY, USA \\
  \texttt{nima@ee.columbia.edu} \\
}

\begin{document}

\maketitle

\begin{abstract}

In this study, we propose a deep neural network for reconstructing intelligible speech from silent lip movement videos. We use auditory spectrogram as spectral representation of speech and its corresponding sound generation method resulting in a more natural sounding reconstructed speech. Our proposed network consists of an autoencoder to extract bottleneck features from the auditory spectrogram which is then used as target to our main lip reading network comprising of CNN, LSTM and fully connected layers. Our experiments show that the autoencoder is able to reconstruct the original auditory spectrogram with a 98\% correlation and also improves the quality of reconstructed speech from the main lip reading network. Our model, trained jointly on different speakers is able to extract individual speaker characteristics and gives promising results of reconstructing intelligible speech with  superior word recognition accuracy. 

\end{abstract}

\section{Introduction}
A phoneme is the smallest detectable unit of a (spoken) language and is produced by a combination of movements of the lips, teeth and tongue of the speaker. However, some of these phonemes are produced from within the mouth and throat and thus, cannot be detected by just looking at a speaker's lips. It is for this reason that the number of visually distinctive units or visemes is much smaller than the number of phoneme making lip reading an inherently difficult task. For example, it is nearly impossible to visually distinguish 's' from 'z' as they are both pronounced with tongue near the gums and differ by the presence of voicing or vibration of the vocal cords in one (in the case of 'z') as opposed to the other (here, 's'). 

In the past, most works have viewed lip reading as a video to text task where the final output is a set of textual sentences corresponding to the lip movements of the speaker. We go one step further by modeling it as a video to speech task where our output is the acoustic speech signal allowing us to recover not just the information but also  the style of articulation. 

This opens an entirely new world of applications in the audio-visual domain - improving audio in existing videos in the sequences where someone is talking, for example blogging videos or news anchoring videos, enabling video-chatting in silent areas like libraries or in noisy environments. 

The challenge of speech generation suggested in related study \cite{}\cite{} is that speech models often lose pitch information and consequently have low intelligibility in the reconstructed speech. In this paper, we propose an end-to-end learning network to 
learn the high dimensional auditory spectrogram with its corresponding audio-generation process  \citet{chi2005multiresolution}. Compared with classical models including LPC, LSP, 
traditional spectrogram, auditory spectrogram preserves both the pitch and resonance information of audio signals, by a 128 dimensional feature at every time step. If the audio is of 8kHz, 
auditory spectrogram will yield to millions of parameters. Moreover, these  auditory spectrogram are highly correlated, which makes it is very difficult for the network to learn the auditory spectrogram directly.

In this paper, we developed two techniques which work well for these tasks. First, we train a deep autoencoder network with additive Gaussian noise, which can robustly recover
original spectrogram  with a low dimensional sparse feature vector even  in noise. Second, we employ a new cost  as a combination of mean squared loss and correlation loss which preserves both the shape and relative amplitude of these audio features.

Based on the deep autoencoder of audio spectrogram, we build a CNN + LSTM network to construct the speech for lip videos. This allows us to generate an acoustic signal that is much more natural sounding than the previous approaches.  Our experiments show that the original speech signal can be reconstructed from the bottleneck features of an auditory spectrogram with a 99.3\% correlation and without any significant loss to quality.  We believe that this sound generation model with auditory spectrogram representation is superior than traditional spectrogram magnitude based models, and has a lot of potentials for future applications.

\section{Related Works}
Lip reading has traditionally been posed as a classification task where words or short phrases from a limited dictionary are classified based on features extracted from lip movements. Some of the early works such as by \citet{ngiam2011multimodal}, \citet{petridis2016deep} and \citet{noda2015audio} used a combination of deep learning and hand-crafted features in the first stage followed by a classifier.  More recently there has been a surge in end-to-end deep learning approaches for lip reading. \citet{wand2016lipreading}, \citet{assael2016lipnet} and  
\citet{chung2016lip} focused on either word level or sentence-level prediction using a combination of convolutional and recurrent networks. 

Our proposed network also follows a similar structure in the sense that the initial layers consist of convolutional layers to extract features from the video followed by an LSTM to model the temporal dependencies. However, we model our output as a generative task over the audio frequency space to directly produce the corresponding speech signal at every time-step. We describe here two works that are most closest to ours. 

 \citet{milner2015reconstructing} reconstructed audio from video, by estimating the spectral envelope using a deep-learning network composed solely of fully connected layers and trained on hand-engineered visual features obtained from mouth region. This approach had the limitation of missing certain speech components such as fundamental frequency and aperiodicity which was then determined artificially thereby compromising quality in order to maximize intelligibility. \citet{ephrat2017vid2speech} modified this technique by using an end-to-end CNN to extract visual features from the entire face while applying a similar approach for modeling audio features using 8th order Linear Predictive Coding (LPC) analysis followed by Line Spectrum Pairs (LSP) decomposition. However, it also suffered from the same missing excitation parameters resulting in an unnatural sounding voice. In a different setting from lip reading, \citet{owens2016visually} also synthesized sound from silent videos of people hitting and prodding objects with a drumstick. They used a CNN to compute features from spacetime images which can be thought of as 3D video CNNs which were fed to a recurrent neural network (RNN). The output of this network was a 10 dimensional representation of the original 42-dimensional spectral envelope projected using PCA.

Our sound generation model differs from these as we use a specific spectrogram inspired by human auditory cortex whose re-synthesis to speech is of a higher quality than traditional spectrograms. This allows us to generate an acoustic signal that is much more natural sounding than the previous approaches. However, the spectrograms in itself are highly correlated and usually difficult for the networks to learn accurately. To bypass this issue, we designed a deep autoencoder to extract compressed features of the spectrogram and forms the target for the main lip reading network.
In the sequel, we will explain our proposed methods for this aim and will elaborately evaluate them and compare the results with the best baseline.
\section{Proposed Networks}
\subsection{Data Preparation} 
Data preparation in this study is really important. We first converted each lip movement video to grayscale and normalized it to have zero mean and unit standard deviation. The face region was then extracted from each frame and resized to have dimensions WxH. It was then divided into $\mathrm{K}$ non-overlapping slices each of length $\mathrm{L_v}$. First and second order temporal derivatives at each frame were calculated to form a 4D tensor of shape (3, H, W, $\mathrm{L_v}$), where 3 is the number of time-derivative channels ($0^{\mathrm{th}}, 1^{\mathrm{st}} \mathrm{and } \ 2^{\mathrm{nd}}$ order). The target bottleneck feature vector was also divided into $\mathrm{K}$ slices with length $\mathrm{L_a}$ and no overlap. 

\subsection{Network I: Autoencoder}
The original audio waveform was first downsampled to 8kHz and converted into its auditory spectrogram representation with 128 frequency bins. In order to compress the auditory features, we designed a deep autoencoder network. It consists of an input layer of size 128 followed by a dense layer of 512 neurons to extract enough features from the input. The subsequent hidden layers then follow the typical autoencoder structure with the number of hidden units decreasing initially from 128 to 64 until the bottleneck layer of size 32 (which was found to be most optimal as shown in the experiments). After this, the structure is mirrored on the other side of the bottleneck (decoder part) to have a hidden layer of 64 neurons and then finally an output layer of the same size as the input layer. Thus, both the input and output of the network is the 128 frequency bin auditory spectrogram. In addition to this, the output of the activation of the bottleneck is contaminated with Gaussian noise. Our experiments show that this improves the robustness of the decoder network. All the non-linearities throughout the autoencoder are LeakyRelu [\citet{maas2013rectifier}] except the bottleneck for which we use sigmoid.  A summary of the proposed autoencoder can be seen in Table ~\ref{table:network-table}. 

\subsection{Network II: Lip Reading Network}
\begin{figure}[h]
\centering
\includegraphics[width=1.0\textwidth]{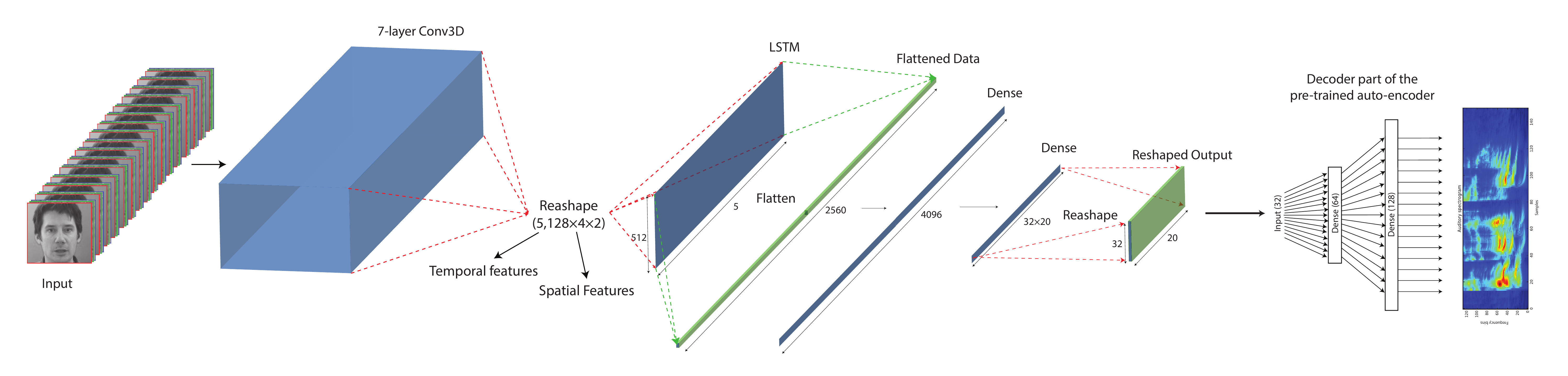}
  \caption{The overall structure for the proposed network. The network gets a video sequence and captures spatiotemporal features and generates coded features of the audio corresponding to that video. Those reconstructed features are then decoded using the pre-trained autoencoder.}
  \label{figure:main-net}
\end{figure}

The input to the main lip reading network is the pre-processed video slice reshaped  as a tensor of shape (None, 3, H, W, $\mathrm{L_{v}}$) where 3 is  the number of time-derivative channels, $\mathrm{H}$ and $\mathrm{W}$ is the height and width of the cropped face images respectively and $\mathrm{L_{v}}$ is the number of video slices. Spatiotemporal features of the video sequence are extracted using a 7-layer 3D convolutional network described in Table ~\ref{table:network-table}. The output of each convolution has the same size as input and is followed by a 3D max pooling layer. All throughout the CNN, the temporal dimensions and order is maintained to enable using an LSTM to model the time-dependencies. Because of this, the output of the convolutional network block is reshaped to a tensor of shape ($\mathrm{L_{v}}$, $\mathrm{N_{f}}$) in which $\mathrm{N_{f}}$ represent the spatial features extracted by the convolutional network. This reshaped tensor is fed into a single-layer LSTM network with 512 units to capture the temporal pattern. Output of this layer is further flattened and fed into a single-layer fully connected network and then finally to the output layer. The output layer has 32$\times\mathrm{L_{a}}$ units to give the 32-bin$\times\mathrm{L_{a}}$-length bottleneck features which is then connected to the decoder part of the pre-trained autoencoder to reconstruct the auditory spectrogram. The overall structure of the proposed network cana be seen in Figure ~\ref{figure:main-net}. 

Non-linearity for all hidden layers of the convolutional network block is LeakyReLU, whereas its output, LSTM and MLP all have ELU [\citet{clevert2015fast}] as the non-linearity. However, the final output layer has sigmoid non-linearity to match the non-linearity at the bottleneck layer of the autoencoder. The audio waveform is reconstructed using the auditory spectrogram toolbox developed by \citet{chi2005multiresolution} (which estimates the phase corresponding to magnitude of the auditory spectrogram and performs the inverse transform to reconstruct the audio waveform).
\begin{table}[h!]
\centering
\begin{scriptsize}
  \caption{Structure of the proposed networks}
  \label{table:network-table}
  \centering
  \begin{tabular}{ccccc}
    \toprule
    \multicolumn{2}{c}{Autoencoder}  &  & \multicolumn{2}{c}{Convolutional network block}  \\
     \cmidrule{1-2} 
      \cmidrule{4-5}
    Layers     & Size  & & Layers     & Size  \\
    \midrule
    Input layer & (None, 128) & &  Input layer & (None, 3, 128, 128, 5)   \\
    \midrule
    Dense (512) &  & & Conv3D (32)  \\
    LeakyReLU    &(None, 512)  & &  LeakyReLU \\
    & & & MaxPooling3D (2,2,1) &  (None, 32, 64, 64, 5)   \\
	\midrule    
    Dense (128) &  & & Conv3D (32)   \\
    LeakyReLU    & (None, 128) & &  LeakyReLU  \\
    & & & MaxPooling3D (2,2,1) &  (None, 32, 32 32, 5)   \\
	\midrule    
    Dense (64) &  & & Conv3D (32)   \\
    LeakyReLU    &  (None, 64) & &  LeakyReLU  \\
    & & & MaxPooling3D (2,2,1) &  (None, 32, 16 16, 5)   \\
	\midrule   
    Dense (32) &   & & Conv3D (64)  \\
    Sigmoid    & (None, 32) & &  LeakyReLU   &  (None, 64, 16, 16, 5)  \\
	\midrule    
    Additive Gaussian Noise &   & & Conv3D (64)  \\
    (Sigma=0.05)   & (None, 32)  & &  LeakyReLU   \\
     & & & MaxPooling3D (2,2,1) &  (None, 64, 8, 8, 5)   \\
	\midrule    
    Dense (64) & & & Conv3D (128)    \\
    LeakyReLU    & (None, 64)  & & LeakyReLU  &  (None, 128, 8, 8, 5) \\
    \midrule    
    Dense (128) & & & Conv3D (128)    \\
    LeakyReLU   & (None, 128) & & ELU (alpha=1.0)   & \\
    & & & MaxPooling3D (2,2,1) &  (None, 128, 4, 4, 5)   \\
    \bottomrule
  \end{tabular}
  \end{scriptsize}
\end{table}
\section{Experiments}
In this section, we elaborate the details about implementation of the network and provide detailed information regarding the evaluation.
\subsection{Dataset}
The dataset used for training the network was the GRID audio-visual corpus [\citet{cooke2006audio}] which consists of audio and video recordings of 34 different speakers (male and female). For each speaker, there are 1000 utterances and each utterance is a combination of six words from a 51-word vocabulary (shown in Table ~\ref{table:grid-table}). Videos and audios are both 3 seconds long with a sampling rate of 25fps and 44kHz respectively. They are pre-processed as described in section 3.1 and then fed to the networks.  We conducted our training and evaluation using videos from two male speakers (S1, S2) and two female speakers (S4, S29).
\begin{table}[h]
\centering
  \caption{GRID vocabulary}
  \label{table:grid-table}
  \centering
  \begin{tabular}{cccccc}
    \toprule
    Command & Color & Preposition & Letter & Digit & Adverb  \\
    \midrule
    bin & blue & at & A-Z & 0-9 & again   \\
    lay & green & by & minus W & & now   \\
    place & red & in & &  & please   \\
    set & white & with & & & soon   \\
    \bottomrule
  \end{tabular}
\end{table}
\subsection{Implementation}
As mentioned before, the pre-processed video was used as input to the network. We fixed the length of each video slice $\mathrm{L_v}$  to 5 (which is equivalent to 200 ms), and length of each audio slice  $\mathrm{L_a}=20$ was also set accordingly such that the number of audio and video slices $\mathrm{K}$ are equal. We found that both width $\mathrm{W}$ and height $\mathrm{H}$ of cropped face images when set to 128 was sufficient to extract enough features. We shuffled all the utterances from all 4 speakers and selected 10 samples for validation, 10 for testing and the rest for training. 

We used Keras [\citet{chollet2015keras}] with Tensorflow backend [\citet{tensorflow2015-whitepaper}] for implementing the network. Initialization of the network weights was performed using the proposed method by \citet{he2015delving}. We used batch normalization [\citet{ioffe2015batch}] for all layers, dropout [\citet{srivastava2014dropout}] of p=0.25 every two layers in convolutional block and L2 penalty multiplier set to 0.0005 for all convolutional layers. For LSTM and MLPs after convolutional blocks, we used dropout of p=0.3 and regularization was not used. We first trained the autoencoder on the 128 frequency bin auditory spectrogram of the training audio samples with a mini-batch size of 128. After training, we extracted the 32-bin bottleneck features which we then provided as target features for the main network. The main lip reading network was trained using a batch size of 32 and the parameter $\alpha$ for ELU non-linearity was set as 1. To improve robustness of the network, we performed data augmentation in each epoch by randomly selecting videos and either flipping them horizontally or adding small Gaussian noise. Optimization was performed using Adam  [\citet{kingma2014adam}] with an initial learning rate of 0.0001, which was reduced by a factor of 5 if validation loss was not improving in 4 consecutive epochs. The loss function we used for all our networks was a combination of mean squared error (MSE) and correlation as given by: 
\begin{equation}
 \lambda \frac{1}{n}\sum_{i}{({y_{i}-\hat{y}_{i}})^{2}} -
\frac{\sum_{i}{(y_{i}-\bar{y})(\hat{y}_{i}-\bar{\hat{y}}) }}{\sqrt{(\sum_{i}{(y_{i}-\bar{y})^{2}}) (\sum_{i}{(\hat{y}_{i}-\bar{\hat{y}})^{2}})}}
\end{equation}
in which,  $\lambda$ is the hyper-parameter for controlling balance between the two loss functions. We show that this loss function  (that we call it CorrMSE ), with  $\lambda$ fixed to 1, performs better than both MSE and correlation.
For auditory spectrogram generation, we used NSRtools, a Matlab toolbox for proposed method by \citet{chi2005multiresolution}. Parameters for all auditory spectrogram generation and audio waveform reconstruction from the spectrogram were $\mathrm{frm\_len}=10,\mathrm{tc}=10, \mathrm{fac}=-2, \mathrm{shft}=-1$. It’s worth mentioning that for extracting better features using the autoencoder, we first compressed the spectrograms by raising them to the power of 1/3 and for retrieving the audio waveform from the output of the decoder we reversed the compression process by raising it to the power of 3 before feeding it to the auditory spectrogram toolbox for waveform reconstruction.

\subsection{Results}
In this section, we first discuss the ability of the autoencoder for coding an auditory spectrogram with 128 frequency bins to N-bin bottleneck features (N<128), then we discuss the necessity of using the autoencoder in the proposed structure and then we study the quality of reconstructed signals. In the end, performance of the proposed method will be compared to the baselines using both human evaluation and quantitative quality measures.

\subsubsection{Quality and accuracy measurement of the reconstructions}
\label{qualmeas}
For evaluating both the autoencoder and the main network, we first reconstruct the auditory spectrogram and measure 2D correlation between the reconstructed ($\hat{\mathbf{Y}}$) and the main spectrogram ($\mathbf{Y}$) using 2D Pearson's correlation coefficient. Then, the audio waveforms corresponding to the spectrograms are recovered using the method in \citet{chi2005multiresolution}, and quality of the reconstructed audio is measured using the standard Perceptual Evaluation of Speech Quality (PESQ), an automated assessment of the speech quality \citet{rix2001perceptual}. Thus, in all evaluations we use an accuracy measure in frequency domain (Corr2D) and a quality measure in time domain (PESQ). We also measure intelligibility of the final reconstructions using Spectro Temporal Modulation Index (STMI) \citet{elhilali2003spectro}.

\subsubsection{Autoencoder}
We trained the autoencoder model on $90\%$ of the spectrograms from the GRID corpus for four speakers, S1 (male), S2 (male), S4 (female) and S29 (female), used $5\%$ for validation during the training and tested it on the remaining $5\%$ which where completely unseen. We used CorrMSE as loss function and the training was done on 50 epochs and validation loss stopped improving after 40 epochs. In our evaluations on the proposed autoencoder, we examine effect of 1. numbers for the bottleneck nodes, 2. additive noise to the bottleneck, 3. dropout on quality of decoded audio. It's worth mentioning that except the study on number of bottleneck nodes, the rest of conditions are with bottleneck\_nodes=32. Table~\ref{table:auto-table} shows the Corr2D and PESQ measures for the mentioned conditions.

\begin{figure}[h]
\centering
\includegraphics[width=\textwidth]{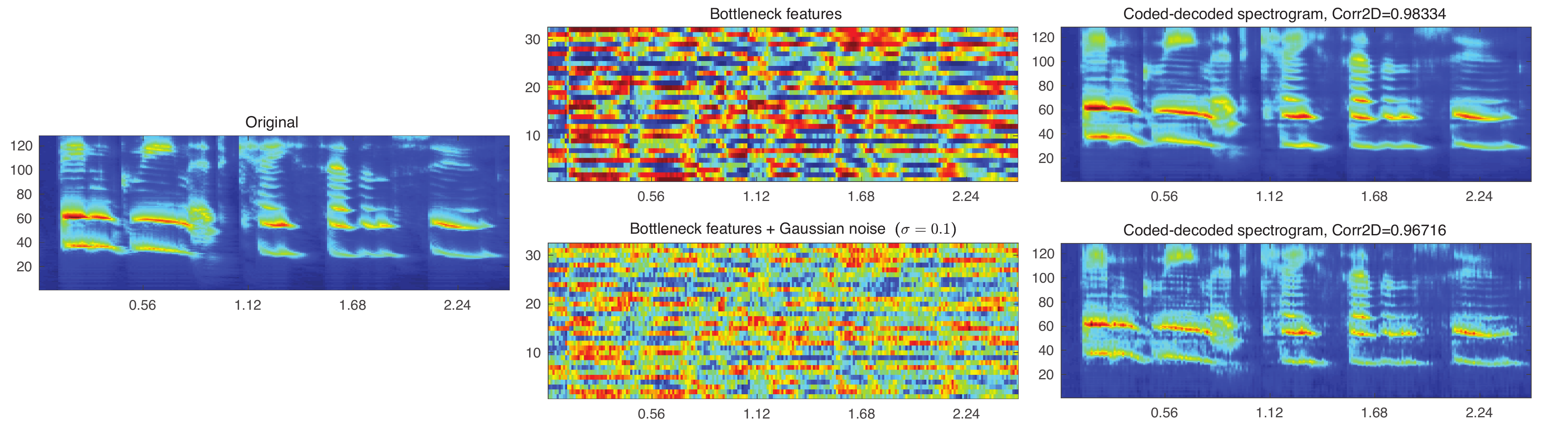}
  \caption{Reconstructed audio spectrograms}
  \label{figure:autoen-sigmoid}
\end{figure}

\begin{table}[h!]
  \caption{Quality and accuracy measures for coded-decoded audio using the proposed autoencoder}
  \label{table:auto-table}
  \centering
  \begin{tabular}{cccccc}
    \toprule
    Measure & 16 nodes & 32 nodes & 64 nodes & No noise (32 nodes) & Dropout (32 nodes) \\
    \midrule
    PESQ & 2.76 & 2.81 & 2.92 & 2.88 & 2.33\\
    Corr2D & 0.98 & 0.98 & 0.99 & 0.97 & 0.95\\
    \bottomrule
  \end{tabular}
\end{table}
It is clear that by increasing number of nodes, both correlation and quality improve. Also, using dropout makes the results worse, and by not using noise we lose the accuracy of reconstruction, however it improves the quality. A visual evaluation of the autoencoder (32 nodes, additive noise) can also be seen in Figure ~\ref{figure:autoen-sigmoid}. As it can be seen, the proposed autoencoder is able to code a spectrogram to 4 times smaller feature space and decode it to the original spectrogram space with a high accuracy even in noisy conditions. This robustness makes it a good choice for being used as target to the main lip-reading network instead of the original spectrogram, since it can handle the unwanted noise or variations in the input videos to the lip-reading network and consequently, variations in the output of it. In section~\ref{necess} we more discuss its effect on the overall result in main lip-reading network.
\subsubsection{Lip-reading network}
\begin{figure}[h]
\centering
\includegraphics[width=\textwidth]{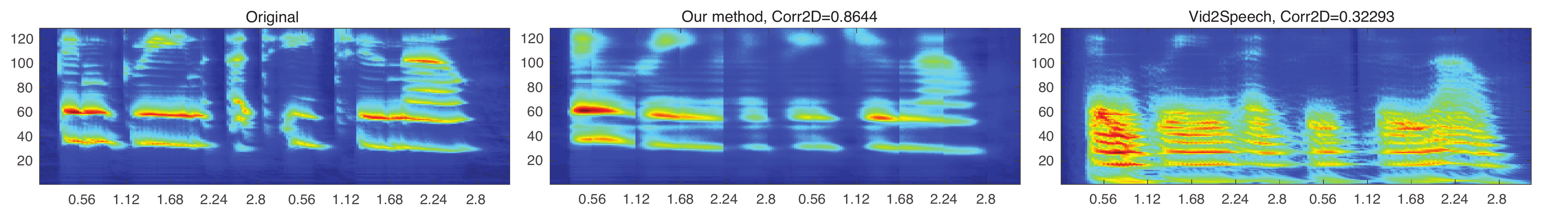}
  \caption{Reconstructed audio spectrograms}
  \label{figure:main-net-results}
\end{figure}

We trained the main lip-reading network with 150 epochs on pre-processed train set (train-validation-test had the same combination as autoencoder's). Validation accuracy stopped improving after around 100 epochs. For a comparison, we also trained-validated-tested the best baseline (up to our knowledge), Vid2Speech \citet{ephrat2017vid2speech}, on the same train-validation-test. We conducted two different class of evaluations on the results of the two methods: 1. Computer-based evaluation, 2. Human evaluation. For the first one we used STMI, PESQ and Corr2D measures discussed in Section~\ref{qualmeas}. Average STMI, PESQ and Corr2D for the two methods can be found in Table~\ref{table:Mquality-table}.
\begin{table}[h!]
  \caption{Quality and accuracy measures for our proposed method compared to Vid2Speech}
  \label{table:Mquality-table}
  \centering
  \begin{tabular}{ccccccc}
    \toprule
    Measure & Method & S1 & S2 & S4 & S29 & Average  \\
    \midrule
    STMI & Our method & \textbf{0.82} & \textbf{0.84} & \textbf{0.84} & \textbf{0.82} & \textbf{0.80}\\
     & Vid2Speech & 0.58 & 0.59 & 0.46 & 0.48 & 0.52\\
    \midrule
    PESQ & Our method & \textbf{2.07} & \textbf{2.01} & 1.61 & \textbf{1.84} & \textbf{1.88}\\
     & Vid2Speech & 1.90 & 1.74 & \textbf{1.79} & 1.62 & 1.76\\
    \midrule
    Corr2D & Our method & \textbf{0.89} & \textbf{0.88} & \textbf{0.88} & \textbf{0.87} & \textbf{0.88}\\
     & Vid2Speech & 0.62 & 0.52 & 0.64 & 0.65 & 0.61\\
    \bottomrule
  \end{tabular}
\end{table}
As it can be seen from the table, the reconstructed audios using the proposed method have both higher quality and intelligibility compared to the baseline. Our method has also been able to get a significantly higher Corr2D compared to Vid2Speech which indicates the high reconstruction accuracy, specifically the pitch information, in frequency space. To shed light on that, we generated spectrograms for a random sample from the test set and compared both reconstructions with the original spectrogram of that sample, which can be seen in Figure ~\ref{figure:main-net-results}.

It is clear from the figure that accuracy of audio reconstruction in frequency space is much more accurate than the baseline. It's also worth mentioning that the model can successfully handle connections between windows and memorizes the time pattern for continuing a phoneme from one window sample to another. Although, both methods fail in retrieving high frequency information which is mainly due to the nature of the task. Not all phonemes can be recovered from lip-movements alone, since valuable information about  speech that is mostly shaped by tongue movements is missing in most of the articulations. For example, although many of the consonants can be recovered solely from lip movements, it is quite hard to reconstruct different vowels accurately.

\subsubsection{Human evaluations}
We conducted a survey on Amazon Mechanical Turk to evaluate the intelligibility and quality of reconstructed speech by our method as well as 'Vid2Speech'. The task was to transcribe each audio sample (human based speech recognition) for analyzing its intelligibility as well as to answer questions for judging the quality of the audio. These questions asked the workers to rate each sample on a scale of 1-5 on the basis of how natural it sounded with 1 being unnatural and 5 being extremely natural. In addition to this, workers also had to guess the gender of the speaker, in which they were given 3 choices: male, female and hard to say. Each audio sample was evaluated by 20 unique workers who were provided with the GRID vocabulary and were allowed to replay the audio unlimited times. Table ~\ref{table:accuracy-table} shows the result of this evaluation.

\begin{table}[h!]
\centering
\begin{footnotesize}
  \caption{Speaker-wise intelligibility \& qualitative assessments for audio-only}
  \label{table:accuracy-table}
  \centering
  \begin{tabular}{c
   ccccccccccc}
  \toprule
   & \multicolumn{5}{c}{\citet{ephrat2017vid2speech}} & & \multicolumn{5}{c}{Ours} \\
   \cmidrule{2-6} \cmidrule{8-12}
   Measure& S1 & S2 & S4 & S29 &Avg. && S1 & S2 & S4 & S29 & Avg.  \\ 
   \midrule
   Accuracy(\%) & 35.2 & 51.2 & \textbf{57.7}  & 59.6 & 50.9 && \textbf{49.3} & \textbf{56.1} & 54.9 & \textbf{63.7} & \textbf{55.8}\\
  Natural sound (1-5) & 1.13 & 1.45 & 1.44 & 1.37 & 1.35 && \textbf{1.69} & \textbf{1.48} & \textbf{1.67} & \textbf{1.67} & \textbf{1.63} \\
Correct Gender (\%) & 58.0 & 77.0 & 21.0 & 17.0 & 43.2 && \textbf{85.83} & \textbf{79.2} & \textbf{83.3} & \textbf{92.0} &   \textbf{85.1} \\
Hard to say (\%) & 36.0 & 16.0 & 42.0 & 32.0 & 31.5 && \textbf{9.16}  & \textbf{12.5} & \textbf{10.0} & \textbf{2.0} & \textbf{8.4} \\
  \bottomrule 
  \end{tabular}
  \end{footnotesize}
\end{table}
We can see that in terms of word accuracy, our model performs significantly better than Vid2Speech for three out of four speakers. This leads us to the conclusion that the reconstructed speech by our method has more intelligibility and improves the word recognition accuracy on the baseline by 5\% when averaged over all four speakers. The accuracy for random guessing is 19\% here. In terms of quality of speech, our method consistently outperforms Vid2Speech. Not only is our reconstructed speech more natural, it also shows significant speaker dependent characteristics such as gender which is due to correct pitch information retrieval. On the other hand, Vid2Speech suffers from lack of pitch information which results in poor speech quality especially for female speakers.

We also summarized the category-wise word accuracies for the two methods and can be found in Table ~\ref{table:word-table}.As expected, the accuracy for 'Letter' is much lower than the other categories. This can be attributed to the fact that alphabets are harder to infer mainly due to the inherent ambiguity in lip-reading of phonemes that get mapped to the same visemes. However, the accuracy for words are significantly higher and they are also better than the performance of Vid2Speech.

\begin{table}[h!]
  \caption{Word category-wise accuracies for audio-only}
  \label{table:word-table}
  \centering
  \begin{tabular}{ccccccc}
    \toprule
    Method & Command & Color & Preposition & Letter &  Digit & Adverb   \\
    \midrule
    \citet{ephrat2017vid2speech} & 58.7\% & 70.6\% & 47.8 \% & \textbf{9.1}\% & \textbf{46.2}\%
    & 73.2\%
    \\ 
    Ours & \textbf{72.7}\% & \textbf{85.7}\% & \textbf{48.8}\% & 6.0\% & 40.0\% & \textbf{81.8}\% \\
    Random selection & 25.0\% & 25.0\% & 25.0\% & 4.0\% & 10.0\% & 25.0 \\
    \bottomrule
  \end{tabular}
\end{table}
\subsubsection{Necessity of using autoencoder and proper loss function}
\label{necess}
As a study on effect of different loss functions on the overall results, we compared results of the main lipreading network trained using MSE, Corr2 and a combination of the two (CorrMSE). We also conducted a test to see how the autoencoder helps the main network to have more accurate reconstructions. Thus, instead of using bottleneck features as the target we directly fed the spectrogram to the output layer of the main lip-reading network. We chose one of the speakers (S29) which had a wide frequency range and subtle high frequency information in its spectrograms, and evaluated the methods in the mentioned conditions on it. Table~\ref{table:quality-table} shows PESQ and Corr2D measures for both using bottleneck features and spectrograms as target to the network using different loss functions.

\begin{table}[h!]
  \caption{Quality measures for different loss functions and different target features}
  \label{table:quality-table}
  \centering
  \begin{tabular}{ccccccc}
    \toprule
    \small Measure & \small MSE(Bott.) & \small Corr(Bott.) & \small CorrMSE(Bott.) & \small MSE(Spec.) & \small Corr(Spec.) & \small CorrMSE(Spec.)   \\
    \midrule
    PESQ & 1.54 & 1.73 & \textbf{1.76} & 1.29 & 1.69 & 1.58\\
    Corr2D & 0.84 & 0.88 & \textbf{0.89} & 0.87 & \textbf{0.89} & 0.88   \\
    \bottomrule
  \end{tabular}
\end{table}
It is clear that using MSE as loss function results in a relatively poor quality of overall reconstructed speech whereas correlation and CorrMSE result in having both higher Corr2D (in frequency) and PESQ (in time) for the speech reconstructions. CorrMSE performs worse than correlation in using spectrogram, however using it as loss function while feeding bottleneck features as targets to the network results in having the best performance among all the conditions. We also examined the effect of three conditions (for the bottleneck nodes) on the overall lip-reading performance: 1. number of bottleneck nodes, 2. feeding noise to the bottleneck nodes while training (the autoencoder), 3. Using dropout while training (the autoencoder). Table~\ref{table:MLquality-table} shows PESQ and Corr2D measures (on reconstructed speech from lip movements) based on the mentioned conditions.

\vspace*{-2mm}
\begin{table}[h!]
  \caption{Quality and accuracy measures while feeding bottleneck features as target}
  \label{table:MLquality-table}
  \centering
  \begin{tabular}{cccccc}
    \toprule
    Autoencoder  & 16 nodes & 32 nodes & 64 nodes & 32 nodes (w/o noise) & 32 nodes (w/ dropout)\\
    \midrule
    PESQ & 1.19 & \textbf{1.76} & 1.26 & 1.09 & 1.29\\
    Corr2D & \textbf{0.89} & \textbf{0.89} & 0.87 & 0.46 & 0.88\\
    \bottomrule
  \end{tabular}
\end{table}
Generally, it's easier for the lip-reading network to reconstruct smaller number of features (as we saw that their loss values were better). But, it's harder for the autoencoder with smaller bottleneck to recover the spectrogram from those features, and it's the opposite for larger bottlenecks. Because of this trade-off, it's clear that there should be a point in the middle for having the balance. It's also clear from the table that using additive Gaussian noise significantly improves overall performance of the lip-reading structure. Even using dropout cannot handle the variations in the output of the main network and it fails in quality. Based on the results in this section, we used 32 nodes bottleneck with additive noise for all our evaluations on the main network.

 \section{Conclusion}
In this paper, we proposed a structure consisting of a deep autoencoder for coding speech and a deep lip-reading network for extracting speech-related features from the face. We showed that such a combination improves both quality and accuracy of the reconstructed audio. We also conducted different tests for comparing our network with a strong baseline and showed that the proposed structure outperforms the baseline in speech reconstruction. Future work is to collect more train data, including emotions, and to propose an end-to-end structure to directly estimate raw waveform from facial speech-related features. Codes and samples for this project are available online on: \href{https://github.com/hassanhub/LipReading}{https://github.com/hassanhub/LipReading}
{\small
\bibliographystyle{apalike}
\bibliography{egbib}
}
\end{document}